\documentclass[letterpaper]{article} 
\usepackage{times}  
\usepackage{helvet}  
\usepackage{courier}  
\usepackage{url}  
\usepackage{graphicx}  
\usepackage{amsmath}

\begin{document}

\title{ACCNet: Actor-Coordinator-Critic Net for ``Learning-to-Communicate'' \\ with Deep Multi-agent Reinforcement Learning}

\author{
Hangyu Mao$^{\scriptscriptstyle 1}$, Zhibo Gong$^{\scriptscriptstyle 2*}$, Yan Ni$^{\scriptscriptstyle 1*}$ and Zhen Xiao$^{\scriptscriptstyle 1}$ \\
$^{\scriptscriptstyle 1}$ Peking University \\
$^{\scriptscriptstyle 2}$ Huawei Technologies Co., Ltd. \\
\thanks{These authors contribute equally to this study.}
}

\maketitle

\begin{abstract}
Communication is a critical factor for the big multi-agent world to stay organized and productive. Typically, most previous multi-agent ``learning-to-communicate'' studies try to predefine the communication protocols or use technologies such as tabular reinforcement learning and evolutionary algorithm, which cannot generalize to the changing environment or large collection of agents directly.

In this paper, we propose an Actor-Coordinator-Critic Net (ACCNet) framework for solving multi-agent ``learning-to-communicate'' problem. The ACCNet naturally combines the powerful actor-critic reinforcement learning technology with deep learning technology. It can learn the communication protocols even from scratch under partially observable environments. We demonstrate that the ACCNet can achieve better results than several baselines under both continuous and discrete action space environments. We also analyse the learned protocols and discuss some design considerations.
\end{abstract}

\section{Introduction}
Communication is an important factor for the big multi-agent world to stay organized and productive. For applications where individual agent has limited capability, it is particularly critical for multiple agents to learn communication protocols to work in a collaborative way, for example: data routing \cite{vicisano1998tcp}, congestion detection \cite{wan2003coda} and air traffic management \cite{agogino2012multiagent}.

However, most previous multi-agent ``learning-to-communicate'' studies try to predefine the communication protocols or use technologies such as tabular reinforcement learning (RL) and evolutionary algorithm, which cannot generalize to the changing environment or large collection of agents directly. We argue that this field requires more in-depth studies with new technologies.

Recently, we researchers have seen the success of Deep MARL, i.e., the combination of deep learning (DL) and multi-agent reinforcement learning (MARL), in many applications, such as self-play Go \cite{Silver2016Mastering}, two-player Pong \cite{tampuu2017multiagent} and multi-player StarCraft \cite{peng2017multiagent}. However, those work either assume full observability of the environment or lack communication among multiple agents.

Naturally, in this paper, we ask and try to answer a question: can we learn multi-agent communication protocols even from scratch under partially observable distributed environments with the help of Deep MARL?

We consider the setting where multiple distributed agents are \emph{fully cooperative} with the same goal to maximize the shared discounted sum of rewards $R$ in a \emph{partially observable} environment. Full cooperation means that all agents receive the same $R$ independent of their contributions. Partially observable environments mean that no agent can observe the underlying Markov states and they must learn effective communication protocols. In fact, the problem setting can be exactly modelled as Dec-POMDP-Com \cite{goldman2003optimizing,goldman2004decentralized}, which is an extension of Dec-POMDP \cite{Oliehoek2012Decentralized,oliehoek2016concise} when considering communication. The novelty is that the communication bandwidth is limited.

The limited communication bandwidth is a common setting for recent ``learning-to-communicate'' studies \cite{Zhang2013Coordinating,chen2016decentralized,sukhbaatar2016learning,foerster2016learning}. Traditional cooperative agents can share sensations, learned policies or even training episodes \cite{tan1993multi,Zhang2013Coordinating}, which is not suitable for real-world applications because communication itself takes up much bandwidth. In our opinion, limited communication bandwidth has two meanings. On the one hand, the message at a specific timestep can be transported using a few packets so that it will not take too much bandwidth. On the other hand, only valuable message is necessary to further reduce the bandwidth requirement. That is to say, message only comes from time to time, and the intermittent time is task-specific. To achieve the former limited bandwidth, we suggest to use deep neural networks to compress the message so that both the message dimension and the packets needed for transporting the message can be controlled. And for the latter, we will introduce corresponding methods based on Gating mechanism and Token mechanism in another paper because of the space limitation.

To this end, we propose an Actor-Coordinator-Critic Net
(ACCNet) framework, which combines the powerful actor-critic RL technology with DL technology. The ACCNet has two paradigms. The first one is AC-CNet, which learns the communication protocols among \emph{actors} with the help of coordinator and keeps critics being independent. However, the actors of AC-CNet inevitably need communication even during execution, which is impractical under some special situations \cite{chen2016decentralized}. The second one is A-CCNet, which learns the communication protocols among \emph{critics} with the help of coordinator and keeps actors being independent. As actors are independent, they can cooperate with each other even without communication after A-CCNet is trained well. Note that, actor and critic are not two different agents but two services in one agent.

We explore the proposed ACCNet under different partially observable environments. Experiments show that: (1) both AC-CNet and A-CCNet can achieve good results for simple multi-agent environments; (2) for complex environments, A-CCNet has a better generalization ability and performs almost like the ideal fully observable models. To the best of our knowledge, this is the first work to investigate multi-agent ``learning-to-communicate'' problem based on deep actor-critic RL architecture under partially observable environment\footnote{One similar work \cite{Lowe2017Multi} from OpenAI is released \emph{at the same time}. Another \emph{concurrent} work \cite{foerster2017counterfactual} from Oxford also uses a similar idea. They do not explicitly address the ``learning-to-communicate'' problem, but we affirm each other's methods and results mutually. A comparison between ACCNet and all those related studies are shown in Table \ref{tab:Comparison}.}.

The rest of this paper starts from a brief review of actor-critic RL algorithms and the releted work. We then present the ACCNet, followed by experiments and conclusion.

\section{Background}
Reinforcement learning (RL) \cite{sutton1998introduction} is a machine learning approach to solve sequential decision making problem. At each timestep $t$, the agent observes a state $s_{t}$ and takes an action $a_{t}$, and then receives a feedback reward $r_{t}$ from the environment and observes a new state $s_{t+1}$. The goal of RL is to learn a policy $\pi(a|s)$, i.e., a mapping from state to action, which can maximize the expected discount cumulative future reward $E[R]=E[\sum_{t=0}^{T}\gamma^{t}r_{t}]$.

Model-free RL algorithms can be divided into three groups \cite{konda2003onactor,grondman2012survey}. (1) Actor-only methods directly learn the parameterized policy $\pi(a|s;\theta)$. They can generate continuous action but suffer from high variance in the estimation of policy gradient. (2) Critic-only methods use low variance temporal difference learning to estimate the Q-value $Q(s,a;w)=E[R;s,a]$. The policy can be derived using greedy action selection, i.e., $\pi(a|s)=a^{*}=\mathop{\arg\max}_{a}Q(s,a;w)$. They are usually used for discrete action as finding $a^{*}$ is computationally intensive in continuous action space. (3) Actor-critic methods jointly learn $\pi(a|s;\theta)$ and $Q(s,a;w)$. They preserve the advantages of both actor-only and critic-only methods.
\begin{figure}[!htb]
    \centering
    \includegraphics[width=5.6cm]{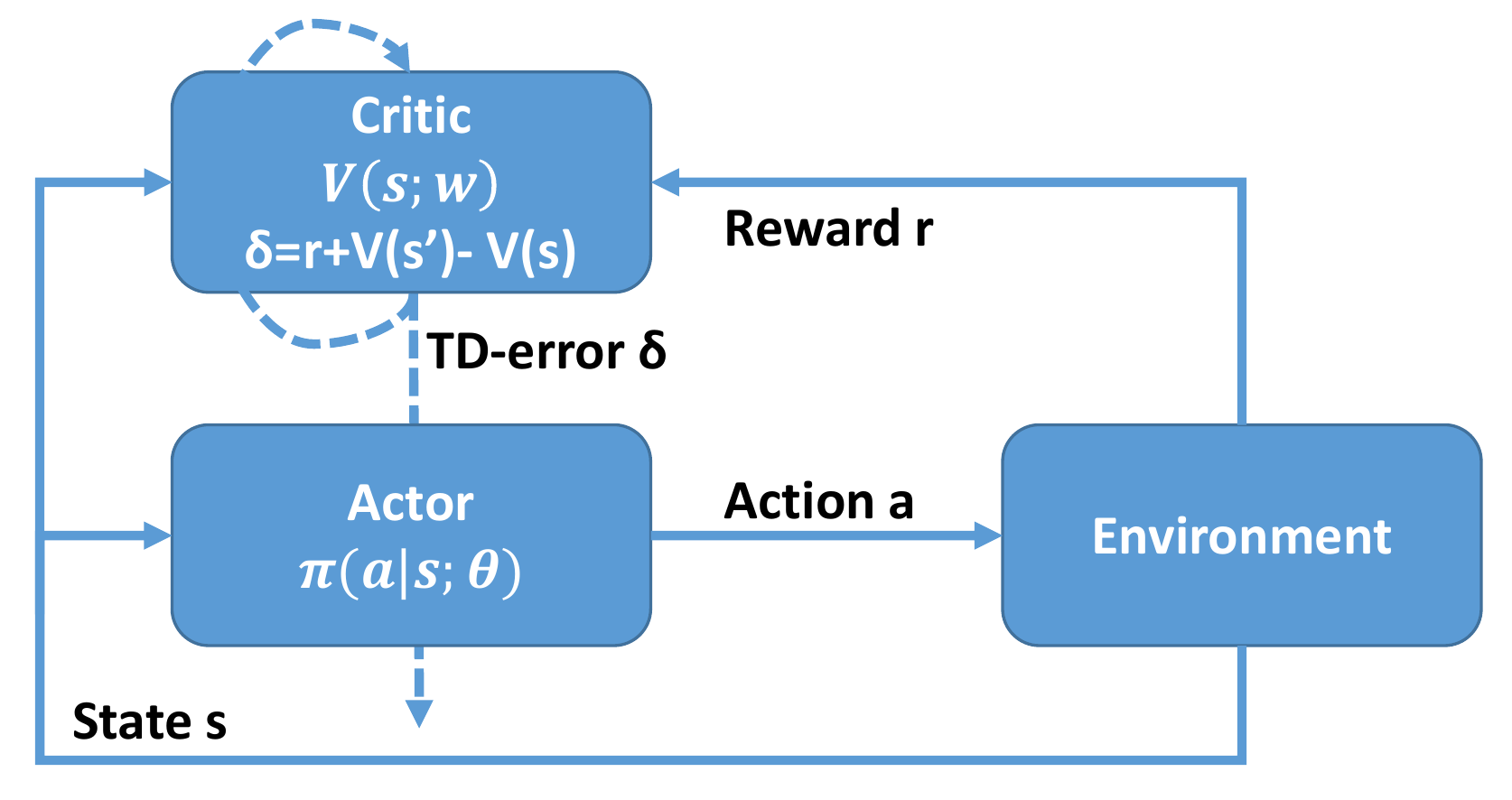}
    \caption{The schematic overview of actor-critic algorithms. The dashed lines indicate that the critic is responsible for updating the actor and itself.}
    \label{fig:AC} 
\end{figure}

The schematic structure of actor-critic methods is shown in Figure \ref{fig:AC}. Two functions reinforce each other: correct actor $\pi(a|s;\theta)$ gives high rewarding trajectory $(s,a,r,s')$, which updates critic $V(s;w)$ or $Q(s,a;w)$ towards the right direction; correct critic $V(s;w)$ or $Q(s,a;w)$ picks out the good action for actor $\pi(a|s;\theta)$ to reinforce. This mutual reinforcement behavior helps actor-critic methods avoid bad local minima and converge faster, in particular for on-policy methods that follow the very recent policy to sample trajectory during training \cite{peng2017multiagent}. Specifically, if actor uses stochastic policy for action selection, the actor and critic are updated based on the following TD-error and Stochastic Policy Gradient Theorem \cite{sutton1998introduction}:
\begin{eqnarray}
    \delta_{t} &=& r_{t}+\gamma V(s_{t+1};w)-V(s_{t};w) \label{equ:SPG1} \\
    \theta_{t+1} &=& \theta_{t} + \alpha*\delta_{t}*\bigtriangledown_{\theta}log\pi(a_{t}|s_{t};\theta) \label{equ:SPG2}
\end{eqnarray}
If actor uses deterministic policy for action selection, they are updated based on the following TD-error and Deterministic Policy Gradient Theorem \cite{silver2014deterministic}:
\begin{eqnarray}
    \delta_{t} &=& r_{t}+\gamma Q(s_{t+1},a_{t+1};w)-Q(s_{t},a_{t};w) \label{equ:DPG1} \\
    \theta_{t+1} &=& \theta_{t} + \alpha*\bigtriangledown_{a}Q(s_{t},a_{t};w)*\bigtriangledown_{\theta}\pi(a_{t}|s_{t};\theta ) \label{equ:DPG2}
\end{eqnarray}
As ACCNet is based on actor-critic methods, the following articles are strongly recommended to read: \cite{Williams1992Simple}, \cite{sutton1998introduction}, \cite{konda2000actor}, \cite{silver2014deterministic} and \cite{lillicrap2015continuous}.

Deep RL (DRL) uses deep neural networks to approximate $\pi(a|s;\theta)$, $Q(s,a;w)$ and/or the environment.

\section{Related Work}
How to learn communication protocols efficiently is critical to the success of multi-agent systems. Most previous work predefine the communication protocols \cite{tan1993multi,Zhang2013Coordinating} and some others use technologies such as tabular RL \cite{kasai2008learning} or evolutionary algorithm \cite{giles2002learning}, which cannot generalize to the changing environment and large collection of agents directly as \cite{sutton1998introduction,foerster2016learning} point out.

Recently, the end-to-end differentiable communication channel embedded in deep neural network has been proven useful for learning communication protocols. Generally, the protocols can be optimized simultaneously while the network is optimized. Our work is an instance of this method, and the most relevant studies include the CommNet \cite{sukhbaatar2016learning}, DIAL \cite{foerster2016learning} and BiCNet \cite{peng2017multiagent}.

CommNet is a single network designed for all agents. The input is the concatenation of current states from all agents. The communication channels are embedded between network layers. Each agent sends its hidden state as communication message to the current layer channel. The averaged message from other agents then is sent to the next layer of a specific agent. However, single network with a communication channel at each layer is not easy to scale up.

DIAL trains a single network for each individual agent. At each timestep, the agent outputs its message as the input of other agents for the next timestep. To learn the communication protocols, it also pushes gradients from one agent to another through the communication channels. However, the message is delayed for one timestep and the environment will be non-stationary in multi-agent situation.

Both CommNet and DIAL are based on DQN \cite{mnih2015human} for discrete action. BiCNet is based on actor-critic methods for continuous action. It uses bi-directional recurrent neural networks as the communication channels. This approach allows single agent to maintain its own internal state and share information with other collaborators at the same time. However, it assumes that agents can know the global Markov states of the environment, which is no so realistic except for some game environments.

Other relevant excellent studies include but not limited to \cite{mordatch2017emergence,das2017learning,havrylov2017emergence}. Those researchers have verified the possibility of learning communication protocols among agents. Nevertheless, we aim at providing a general framework to ease the learning of communication protocols among agents.

\section{Actor-Coordinator-Critic Net Framework}
In this section, we present two paradigms of ACCNet framework for learning communication protocols based on actor-critic models.
\begin{figure*}[!htb]
    \centering
    \includegraphics[width=12cm]{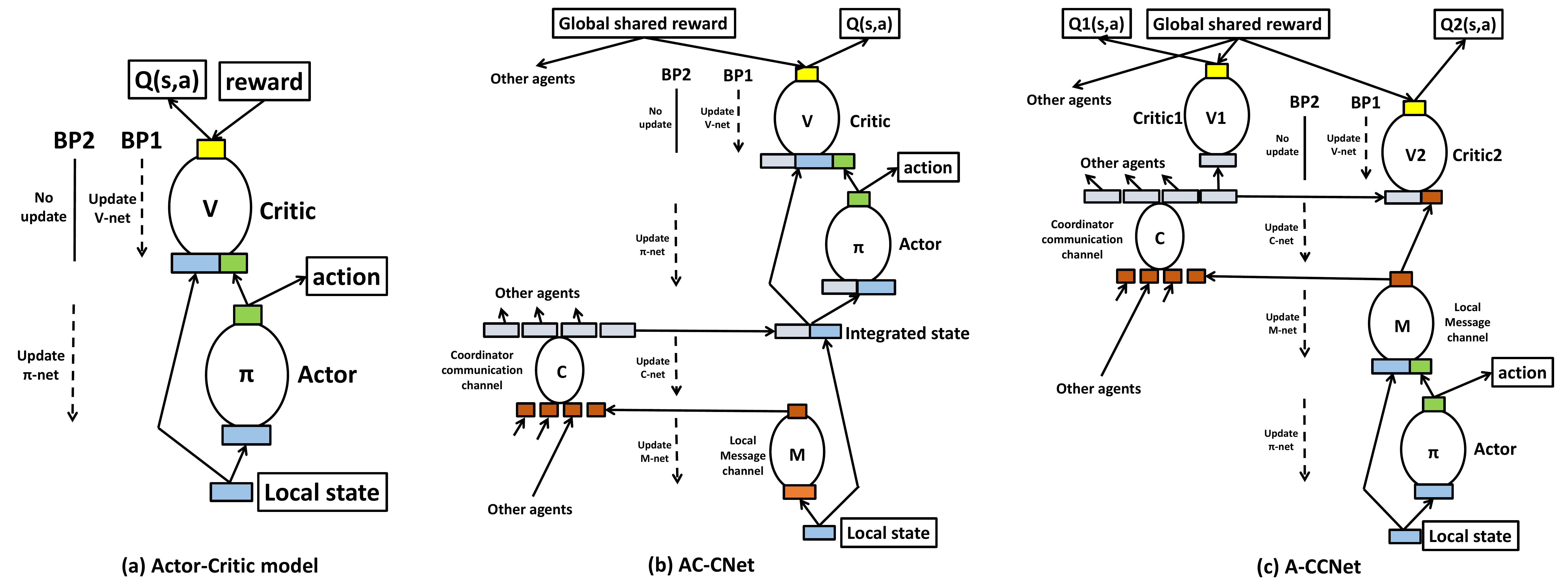}
    \caption{The proposed ACCNet.}
    \label{fig:DeepACCNet} 
\end{figure*}
\subsection{AC-CNet}
The most straightforward approach is to build a communication channel between actors and keep critics being independent. As shown in Figure \ref{fig:DeepACCNet}(b), a coordinator communication channel is used for \emph{coordinating the actors} to generate coordinated actions, so we call this paradigm AC-CNet. Specifically, each agent encodes its local state into a local message and sends it to the coordinator, which further generates the global communication signal for this agent considering messages from all other agents. As the global signal is an encoding of all local messages, we expect that it can catch the global information of the system. The integrated state is the concatenation of local state and global signal, which will be fed as input into the actor-critic model. Then the whole AC-CNet is trained as the original actor-critic model.

However, the AC-CNet inevitably needs communication between actor and coordinator to get the global information even during execution, which is impractical under some special situations \cite{chen2016decentralized,dobbe2017fully}.

\subsection{A-CCNet}
Can those agents generate actions as if they have shared the global knowledge even without communication during training? What about during execution? The answer may be NO at the first glance and we also think so. Fortunately, machine learning has a fascinating property that we can do prediction after one model is trained and the auxiliary data on which the model is trained need no longer to be kept. We ask ourselves that can we move the communication among actors into critics so that actors can independently take actions according to their specific states during execution and the auxiliary critics during training need no longer to be kept. In fact, it is possible for actor-critic methods. However, both actor-only and critic-only methods are unsuitable for this task because the training and execution mechanisms of these methods are exactly the same.

As shown in Figure \ref{fig:DeepACCNet}(c), a coordinator communication channel is used for \emph{coordinating the critics} to generate better estimated Q-values, so we call this paradigm A-CCNet. Specifically, the actor in A-CCNet is the same as the actor in the original actor-critic model shown in Figure \ref{fig:DeepACCNet}(a), but the critics should communicate with each other through coordinator before they can generate the estimated Q-values. Compared to AC-CNet where communication occurs among actors and the communication signal can only encode local state, A-CCNet put communication among critics where both state and action can be encoded into the communication signal. So we expect that A-CCNet can generate better policies, which has been confirmed by the experiments.

Besides, there are two designs for the critic. Critic1 uses the global signal to generate Q-values directly, while critic2 combines global signal and local message to generate Q-values. For both of the two designs, actors can generate their actions independently without communication during execution.

\subsection{Formal Formulation of ACCNet}
For AC-CNet, as critics are independent, we can update each agent based on Equation (\ref{equ:SPG1}-\ref{equ:DPG2}) just like updating single actor-critic agent. One key difference is that we need to push the gradients of actors into the coordinator communication channels so that the communication protocols can also be optimized simultaneously.

For A-CCNet, as critics communicate with each other, the critic network of the $i$-th agent is now $V^{i}(s^{i},s^{g};w^{i})$ or $Q^{i}(s^{i},a^{i},s^{g};w^{i})$, where $s^{g}$=$f(s^{1},...,s^{N},a^{1},...,a^{N})$ is the global communication signal\footnote{Generally speaking, $f$ is a injective function. Besides, using $V^{i}(s^{i},s^{g};w^{i})$ and $Q^{i}(s^{i},a^{i},s^{g};w^{i})$ for discrete and continuous action separately is natural. However, for discrete action, $s^{g}$ is a function of only the states $(s^{1},...,s^{N})$; without knowing the actions $(a^{1},...,a^{N})$, Equation (11) can no longer be true.}. We can then extend Equation (\ref{equ:SPG1}-\ref{equ:DPG2}) into multi-agent formulations:
\begin{eqnarray}
    \delta^{i}_{t} &=& r_{t}+\gamma V^{i}(s^{i}_{t+1},s^{g}_{t+1};w^{i})-V^{i}(s^{i}_{t},s^{g}_{t};w^{i}) \label{equ:mSPG1} \\
    \theta^{i}_{t+1} &=& \theta^{i}_{t} + \alpha*\delta^{i}_{t}*\bigtriangledown_{\theta^{i}}log\pi^{i}(a^{i}_{t}|s^{i}_{t};\theta^{i}) \label{equ:mSPG2} \\
    y^{i}_{t} &=& r_{t}+\gamma Q^{i}(s^{i}_{t+1},a^{i}_{t+1},s^{g}_{t+1};w^{i}) \\
    \delta^{i}_{t} &=& y^{i}_{t}-Q^{i}(s^{i}_{t},a^{i}_{t},s^{g}_{t};w^{i}) \label{equ:mDPG1} \\
    \bigtriangledown_{\theta^{i}_{t}} &=& \bigtriangledown_{a^{i}}Q^{i}(s^{i}_{t},a^{i}_{t},s^{g}_{t};w^{i})*\bigtriangledown_{\theta^{i}}\pi^{i}(a^{i}_{t}|s^{i}_{t};\theta^{i}) \\
    \theta^{i}_{t+1} &=& \theta^{i}_{t} + \alpha*\bigtriangledown_{\theta^{i}_{t}} \label{equ:mDPG2}
\end{eqnarray}

Our primary insight about ACCNet (especially A-CCNet) is that once each agent knows the states and actions from other agents, the environment could be treated stationary regardless of the changing policies. More formally, Equation (11) always keeps true for any agent indexed by $i$ with any changing policies $\pi^{i} \neq \pi^{'i}$ \cite{Lowe2017Multi}:
\begin{eqnarray}
\begin{aligned}
P(s^{i}_{t+1}|s^{i}_{t};Env) =& P(s^{i}_{t+1}|s^{i}_{t};s^{g}_{t},\pi^{1},...,\pi^{N})
\\ =&
P(s^{i}_{t+1}|s^{i}_{t};s^{g}_{t})
\\ =&
P(s^{i}_{t+1}|s^{i}_{t};s^{g}_{t},\pi^{'1},...,\pi^{'N})
\end{aligned}
\end{eqnarray}

\subsection{Some Comparisons}
Before the comparison, we first introduce the two \emph{concurrent} studies mentioned in Footnote 1, i.e., COMA \cite{foerster2017counterfactual} from Oxford and MADDPG \cite{Lowe2017Multi} from OpenAI.

COMA, MADDPG and A-CCNet share a similar idea: accelerating training with the help of critics and executing in real environment only based on actors. But the research purposes are different. COMA aims at solving the credit assignment problem in multi-agent cooperative environments. MADDPG wants to investigate both cooperation and competition among agents. The proposed ACCNet tries to provide a general framework to ease the learning of communication protocols among agents even from scratch. Specifically, COMA is based on Stochastic Policy Gradient Theorem \cite{sutton1998introduction} and REINFORCE \cite{Williams1992Simple} algorithm. It uses a counterfactual baseline and a centralised critic to address multi-agent credit assignment problem. However, they only do experiments for discrete action space environments and assume that the critic can get the entire game screen. MADDPG extends DDPG \cite{silver2014deterministic,lillicrap2015continuous} into multi-agent environments. The authors verify that this method is suitable for both cooperative and competitive tasks. However, their experiments are limited to continuous action space environments.

As COMA and MADDPG do not address the ``learning-to-communicate'' problem explicitly, both of them use the states and actions of all other agents directly, without considering the communication cost. Nevertheless, we affirm each other's methods and results mutually.

Now, we are ready to give a brief comparison as shown in Table \ref{tab:Comparison}. As we can see, ACCNet has a better adaptability for different situations.
\begin{table}[!htb] 
    \setlength{\tabcolsep}{3.2pt} 
    \caption{\label{tab:Comparison} Comparisons between ACCNet and related work. M1, M2, M3, M4 and M5 stand for CommNet, DIAL, BiCNet, MADDPG and COMA separately.}
    \begin{center}
    \begin{tabular}{|l|c|c|c|c|c|c|} 
        \hline 
        & \bf M1 & \bf M2 & \bf M3 & \bf M4 & \bf M5 & \bf ACCNet \\
        \hline
        fully cooperative & Y & Y & Y & Y & Y & Y \\
        \hline
        discrete action  & Y & Y & Y & N & Y & Y \\
        \hline
        continuous action & N & N & Y & Y & N & Y \\
        \hline
        parti. observable & Y & Y & N & Y & N & Y \\
        \hline
        distri. agents & N & Y & N & Y & N & Y \\
        \hline
        limited bandwith & N & Y & N & N & N & Y \\
        \hline
        indep. execution & N & N & N & Y & Y & Y \\
        \hline
        \hline
        \multicolumn{7}{|l|}{M4 can also deal with competitive tasks very well.} \\
        \hline
        \multicolumn{7}{|l|}{M5 can address the credit assignment problem very well.} \\
        \hline
    \end{tabular}
    \end{center}
\end{table}

\section{Experiments}
In this section, we test the proposed ACCNet under both continuous and discrete action space environments. Those environments are partial observable with multiple distributed and fully cooperative agents. 

\subsection{Continuous Action Space Environment}
\begin{figure}[!htb]
    \centering
    \includegraphics[width=7.6cm,height=2.6cm]{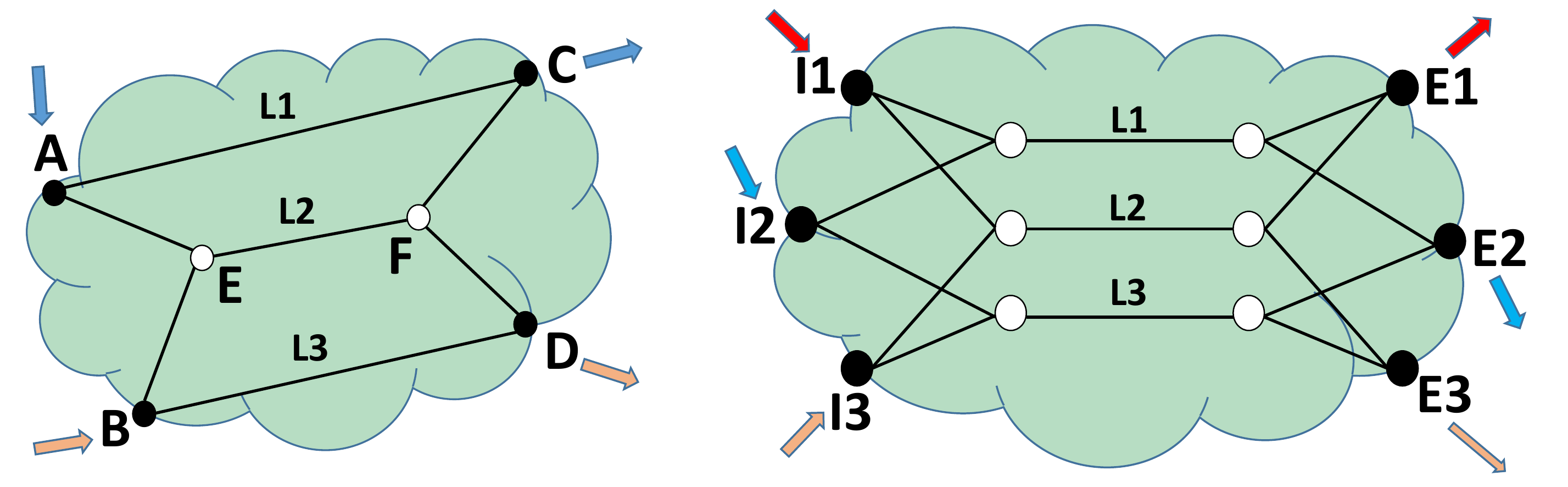}
    \caption{TwoIE and ThreeIE topologies for network flow control studies \cite{Elwalid2001MATE,Kandula:2005:WTR:1080091.1080122}. Link L1/L2/L3 are bottleneck links in both topologies.}
    \label{fig:two-threeIE} 
\end{figure}

\textbf{Problem Definition.}
For continuous action space environment, we focus on the Network Routing Domain problem modified from \cite{ruadulescu2017analysing}. Currently, the Internet is made up of many ISP networks. In each ISP network, as shown in Figure \ref{fig:two-threeIE}, there are several edge routers. Two edge routers are combined as ingress-egress router pair (IE-pair). The $i$-th IE-pair has a input flow demand $F_{i}$ and $K$ available paths that can be used to deliver the flow from ingress-router to egress-router. Each path $P^{k}_{i}$ is made up of several links and each link can belong to several paths. The $l$-th link $L_{l}$ has a flow transmission capacity $C_{l}$ and a link utilization ratio $U_{l}$. As we know, high link utilization ratio is bad for dealing with burst traffic, so we want to find a good traffic splitting policy jointly for all IE-pairs across their available paths to minimize the maximum link utilization ratio in the network.
\begin{figure}[!htb]
    \centering
    \includegraphics[width=6.5cm,height=2.6cm]{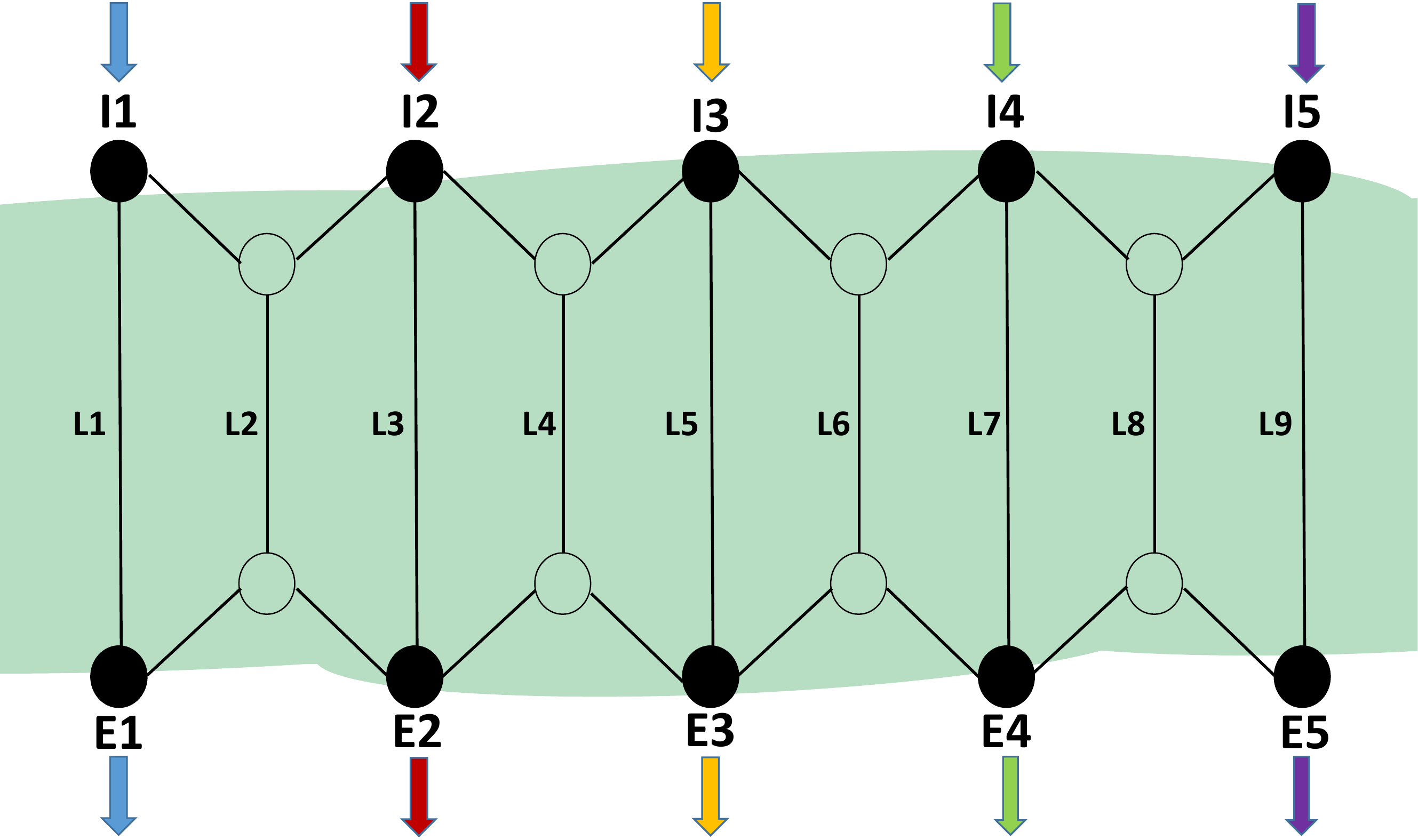}
    \caption{FiveIE network topology for scalability test. Link L1$\sim$L9 are bottleneck links.}
    \label{fig:fiveIE} 
\end{figure}

\textbf{Setting.} We design the following RL elements.

\emph{State.} Current traffic demand and static network topology information are available. We also encode the estimated link utilization ratio into the state. Specifically, the local state is $s=[F_{i}, U^{l}_{i}, max(0, 1-U^{l}_{i}), max(0, U^{l}_{i}-1)]$. 

\emph{Action.} The ingress-router should generate a splitting ratio $y^{k}_{i}$ with a constraint $\sum_{k}y^{k}_{i}=1$ for current traffic demand $F_{i}$. So the softmax activation is chosen as the final layer of actor network. This design is natural for the continuous action with sum-to-one constraint. 

\emph{Reward.} As we want to minimize the maximum link utilization ratio, we set the reward signal to $r=1-max(U_{l})$.

\textbf{Baselines.}
As with CommNet and BiCNet, we also use the following baselines.

\emph{Independent controller (IND):} each agent learns its own actor-critic network without any communication.

\emph{Fully-connected controller (FC):} all agents are controlled by a big fully-connected actor-critic network to learn the traffic splitting policy. The communication channel is embedded in the network without any bandwidth limitation.

IND model is the worst situation and FC model can be seen as the ideal situation. Besides, as mentioned before, we design two kinds of critics for A-CCNet: all critics share the same Q(s,a), or each critic separately learns its own Q(s,a). So we have the following models: IND, FC-sep, FC-sha, AC-CNet, A-CCNet-sep and A-CCNet-sha.

\textbf{Experiment Results.} In this environment, we care about \emph{convergence ratio (CR)} of all independent experiments and \emph{maximum link utilization ratio (MLUi)} of the i-th bottleneck link after convergence. All results are shown in Table \ref{tab:resultTwoIE} and \ref{tab:resultFiveIE}. Due to space limitation, we put the results of ThreeIE in the supplementary material.

As we can see, all models have high \emph{CR} and low \emph{MLUi} for simple TwoIE topology. But A-CCNet has a better performance than AC-CNet and IND. It even has a similar performance with the ideal fully observable FC model. For complex FiveIE topology, the performances of AC-CNet and IND drop severely, while A-CCNet can still keep its ability of performing almost like the ideal FC model. The reason may be that A-CCNet has more global information than other models (except for FC model): A-CCNet put communication among critics where both local state and action can be encoded into the communication signal while the communication signal of AC-CNet can only encode local state and IND does not exchange information at all. In this case, more information means that the environment could be seen stationary as illustrated by Equation (11).
\begin{table}[!htb] 
    \caption{\label{tab:resultTwoIE} The CR and MLUi of TwoIE topology. Results are averaged over 30 independent experiments.}
    \begin{center}
    \begin{tabular}{|l|c|c|c|c|} 
        \hline 
        & \bf CR& \bf MLU1 & \bf MLU2 & \bf MLU3 \\
        \hline
        IND & 0.655 & 0.713 & 0.724 & 0.716 \\
        \hline
        FC-sep & \bf 0.967 & \bf 0.707 & \bf 0.704 & \bf 0.709 \\
        \hline
        FC-sha & 0.967 & 0.710 & 0.702 & 0.715 \\
        \hline
        \hline
        AC-CNet & 0.433 & 0.712 & 0.713 & 0.733 \\
        \hline
        A-CCNet-sep & \bf 0.9 & \bf 0.708 & \bf 0.698 & \bf 0.714 \\
        \hline
        A-CCNet-sha & 0.9 & 0.734 & 0.707 & 0.718 \\
        \hline
    \end{tabular}
    \end{center}
\end{table}
\begin{table}[!htb] 
    \setlength{\tabcolsep}{3.8pt} 
    \caption{\label{tab:resultFiveIE} The CR and MLUi of FiveIE topology. Results are averaged over 30 independent experiments.}
    \begin{center}
    \begin{tabular}{|l|c|c|c|c|c|} 
        \hline 
        & \bf CR& \bf MLU2 & \bf MLU4 & \bf MLU6 & \bf MLU8 \\
        \hline
        IND & 0.1 & 0.817 & 0.879 & 0.891 & 0.828 \\
        \hline
        FC-sep & \bf 0.8 & 0.818 & 0.8 & \bf 0.797 & \bf 0.822 \\
        \hline
        FC-sha & 0.767 & \bf 0.817 & \bf 0.767 & 0.836 & 0.835 \\
        \hline
        \hline
        AC-CNet & 0.0 & - & - & - & - \\
        \hline
        A-CCNet-sep & \bf 0.567 & \bf 0.751 & \bf 0.809 & \bf 0.800 & \bf 0.799 \\
        \hline
        A-CCNet-sha & 0.467 & 0.799 & 0.810 & 0.810  & 0.805 \\
        \hline
    \end{tabular}
    \end{center}
\end{table}

\begin{figure}[!htb]
    \centering
    \includegraphics[width=6.6cm]{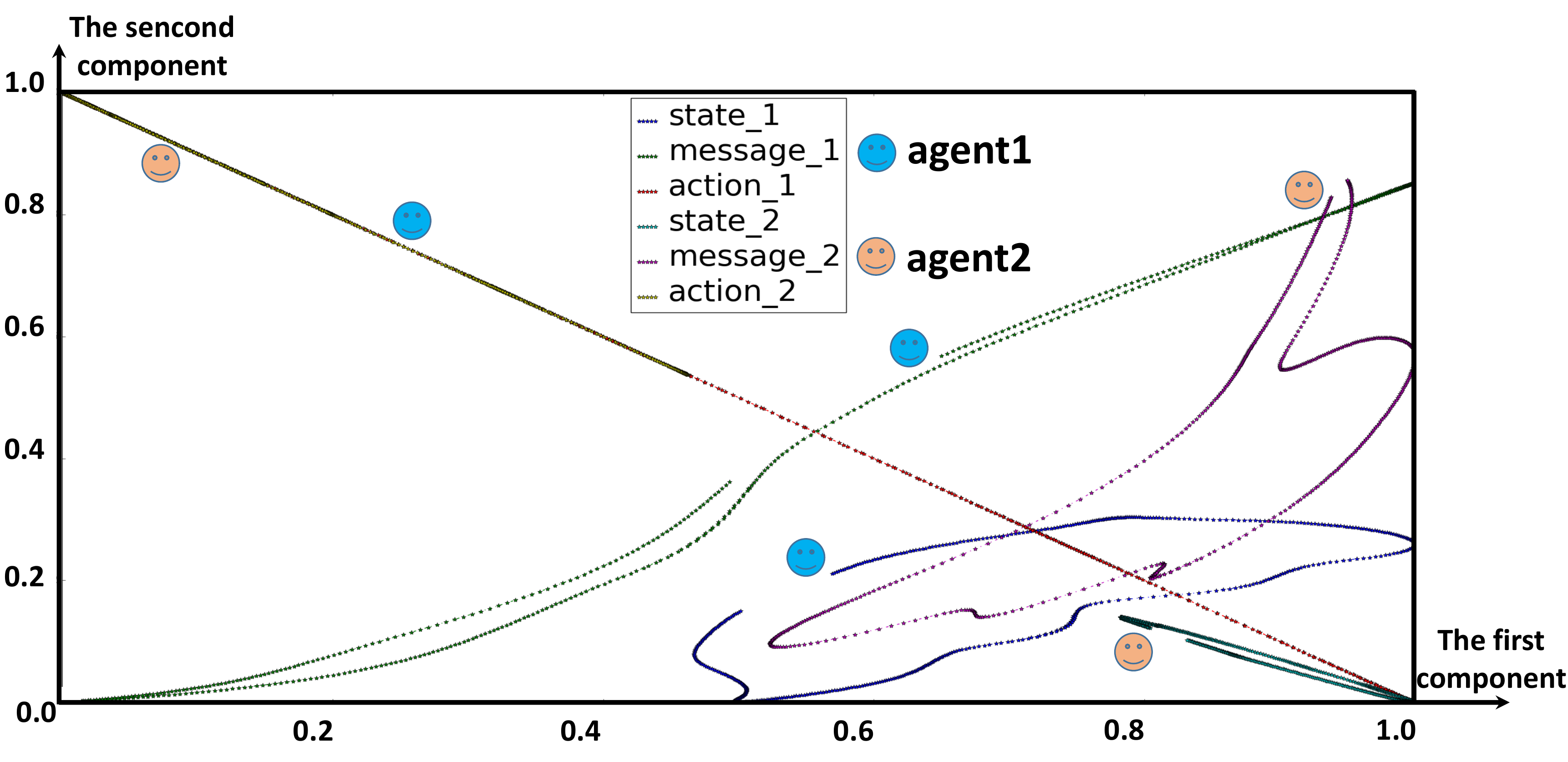}
    \caption{The state-message-action changing of one convergent experiment on topology TwoIE with model AC-CNet. Only the 2D PCA projections of the original data are shown as done in CommNet.}
    \label{fig:twoIE-StateMessageAction} 
\end{figure}

\textbf{Communication Message Analysis.} We show the state-message-action changing of one convergent experiment in Figure \ref{fig:twoIE-StateMessageAction}. As the value of state become large (for example, more packets should be transmitted), agent1 will emit large message value while agent2 usually emits small message value. For action value, if agent1 splits more traffic to L1, agent2 will split more traffic to L2 because L2 is now underused. Besides, agent1 has a wider range of state value, so the message value and action value generated by agent1 are also wider than agent2. Those sophisticated and coordinated behaviors are critical for MARL systems to stay organized.

\subsection{Discrete Action Space Environment}
\begin{figure}[!htb]
    \centering
    \includegraphics[width=4.5cm]{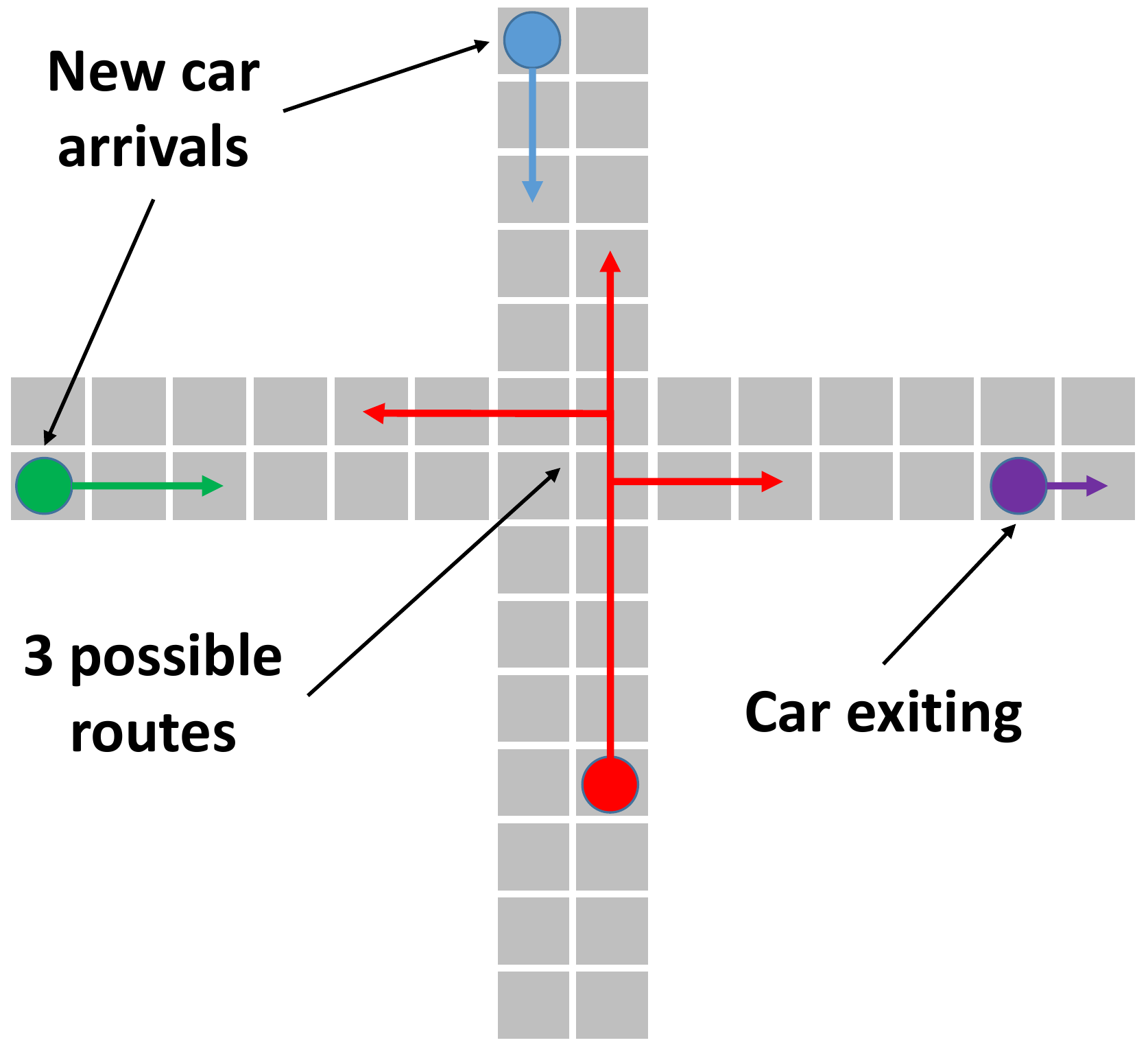}
    \caption{The environment of traffic junction task.}
    \label{fig:TrafficJunction} 
\end{figure}

\textbf{Problem Definition.}
We consider the Traffic Junction problem modified from \cite{sukhbaatar2015mazebase,sukhbaatar2016learning}. As shown in Figure \ref{fig:TrafficJunction}, four cars are deriving on the 4-way junction road. New car will be generated if one car reaches its destination at the edge of the grid. The simulation will be classified as a failure if location overlaps have occurred in 40 timesteps. Our target is to learn a car driving policy so that we can get low failure rate (FR).

\textbf{Setting.}
We use the same RL elements as in CommNet.

\emph{State.} All cars can only know its location and driving direction. They cannot see other cars. So we represent the local state as a one-hot vector set $\{location, direction\}$.

\emph{Action.} A car has two possible actions: gassing itself by one cell on its route or braking to stay at its current location.

\emph{Reward.}  A collision incurs a reward $r_{coll}$=-10.0, and each car gets reward of $r^{\tau}_{time}$=-0.01$\tau$ at each timestep to discourage a traffic jam, where $\tau$ is the total timesteps since the car arrived. So the total
reward at time $t$ is: $r(t)=C^{t}r_{coll}+\sum_{i=1}^{N^{t}} r^{\tau_{i}}_{time}$, where $C^{t}$ is the number of location overlaps at time $t$, and $N^{t}$ is the number of cars. This setting is the same as CommNet.

\textbf{Experiment Results.} Table \ref{tab:resultTrafficJunction} shows the results of this task. After training the models 300 episodes as CommNet, the proposed A-CCNet can get lower FR than CommNet and other baselines. When the training episode increases to 600, A-CCNet can further get a lower FR and a higher CR, while other models cannot get the same results.
\begin{table}[!htb] 
    \caption{\label{tab:resultTrafficJunction} Averaged results of 30000 experiments on traffic junction task. The results of CommNet and Discrete-CN are directly cited from \cite{sukhbaatar2016learning}. Please note that the two environments have some nuances.}
    \begin{center}
    \begin{tabular}{|l|c|c|c|c|} 
        \hline 
        & \multicolumn{2}{|c|}{\bf 300 episodes} & \multicolumn{2}{|c|}{\bf 600 episodes} \\
        \hline
        & \bf CR & \bf FR (\%) & \bf CR & \bf FR (\%) \\
        \hline
        IND & 0.57 & 11.18 & 0.6 & 18.35 \\
        \hline
        FC-sep & \bf 0.8 & 12.76 & \bf 0.8 & 11.95 \\
        \hline
        FC-sha & 0.73 & \bf 12.69 & 0.76 & \bf 10.05 \\
        \hline
        \hline
        AC-CNet & 0.47 & 12.04 & \bf 0.73 & 14.25 \\
        \hline
        A-CCNet-sep & \bf 0.6 & 10.66 & \bf 0.73 & 10.48 \\
        \hline
        A-CCNet-sha & 0.53 & \bf 7.88 & \bf 0.73 & \bf 4.96 \\
        \hline
        \hline
        CommNet (CN) & - & 10.0 & - & - \\
        \hline
        Discrete-CN & - & 100.0 & - & - \\
        \hline
    \end{tabular}
    \end{center}
\end{table}

\begin{figure}[!htb]
    \centering
    \includegraphics[width=7.6cm]{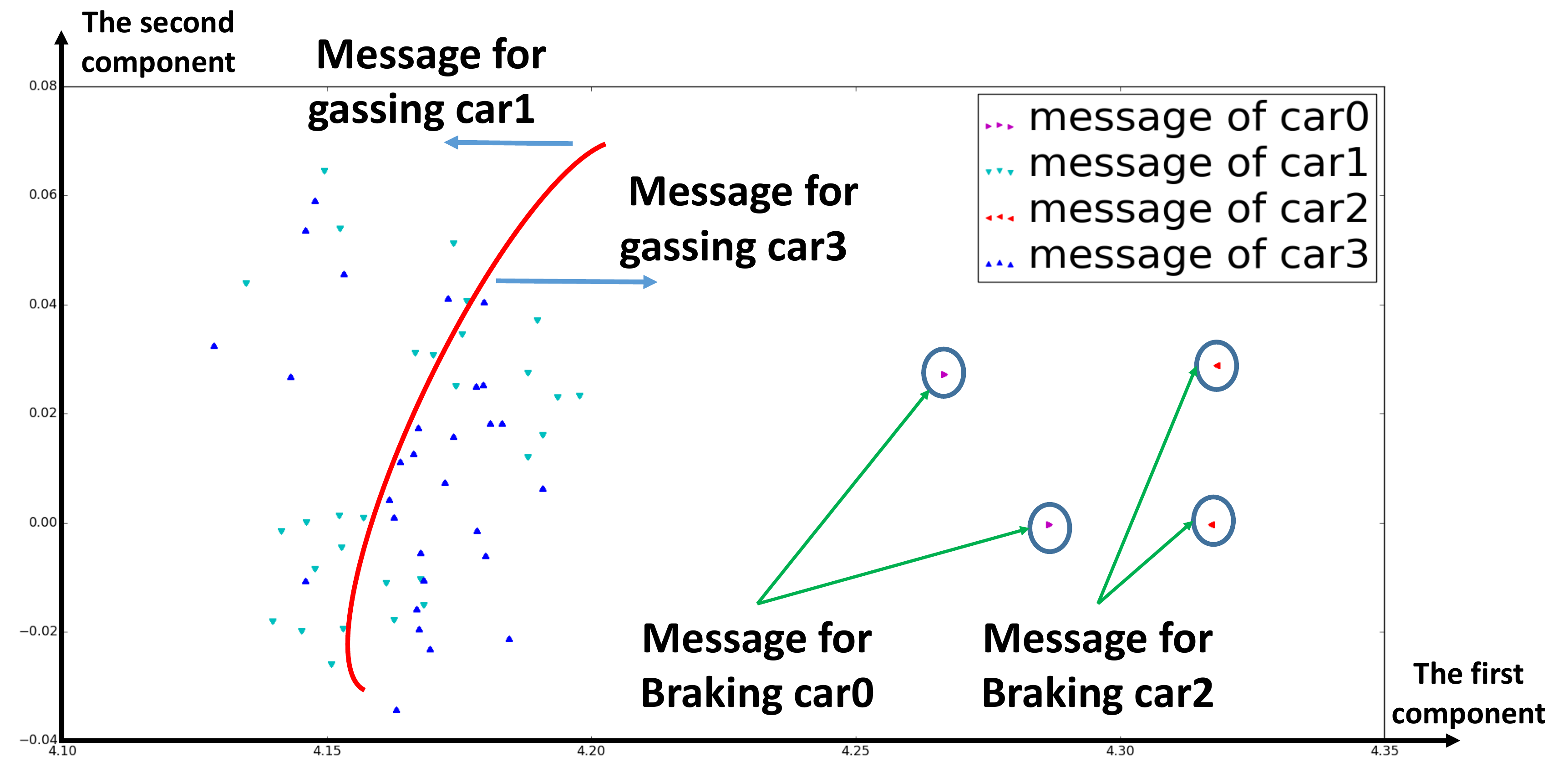}
    \caption{The emitted messages by different cars of one special learned car deriving policy. Only the 2D PCA projections of the original messages are shown.}
    \label{fig:FourCarsMessage} 
\end{figure}

\textbf{Communication Message Analysis.} We find a special car driving policy where the left car0 and the right car2 always brake to make space for the above car1 and the below car3. We illustrate the emitted messages by different cars under this policy in Figure \ref{fig:FourCarsMessage}. As we can see, messages for braking and gassing are naturally separated. For the same type (no matter braking or gassing) of messages, they can also be separated by different cars so that the ACCNet can distinguish them. Besides, gassing message is more diverse than braking message. The reason may be that braking positions are a few (near the junction) while each position of the grid road needs a different gassing message. 

\section{Design Discussion of ACCNet}
As we know, some design choices are very important for the success of DRL in real-world applications. For example, experience replay, frame skipping, target network, reward clipping, asynchronous training, auxiliary task and even methods of DL such as batch normalization, attention mechanism and skip connection are widely adopted \cite{mnih2015human,schaul2015prioritized,mnih2016asynchronous,jaderberg2016reinforcement,lillicrap2015continuous,sorokin2015deep,sukhbaatar2016learning}. In this section, we briefly present a few design choices used by ACCNet, hoping that other researchers can confirm their usefulness for new environments. Note that all those design choices need to be further studied.

(1) The embedded communication channel. We suggest to use deep neural networks to encode the communication message so that the final message dimension is controlled to be independent of the dimension of the original information. And most importantly, as the communication channel is embedded in deep neural networks, the communication protocols can be learned even from scratch in an end-to-end differentiable way while the network is optimized.

(2) The concurrent experience replay (CER). Experience replay is beneficial for single-agent RL. Except for making the collection process of training data more efficient, it can also break the correlation among sequential training data to accelerate the convergence of models.  However, as \cite{foerster2016learning} point out, it is necessary to disable experience replay for MARL due to the non-concurrent property of local experiences when sampled independently for each agent. We propose CER to address this problem. Generally speaking, CER samples the concurrent experiences collected \emph{at the same timestep} for all agents as they are trained concurrently. In fact, \cite{Omidshafiei2017Deep} use the same replay method and name it CER. We refer the readers to this paper for further details.

(3) The current episode experience replay (CEER). Traditional experience replay methods \emph{uniformly} sample a batch of experiences from replay buffer as training examples for model updating. \cite{schaul2015prioritized} introduce prioritized experience replay based on the magnitude of TD-error to accelerate learning. The proposed CEER can be seen as a time-prioritized replay method. CEER keeps all experiences of \emph{current} episode in a temporary buffer and combines them with experiences from the main replay buffer as training examples at the end of each episode. Our preliminary experiments show the effectiveness of this method. Detailed analyses can be found in the supplementary material.

(4) Disabled experience replay for discrete action space environments. We find that the training of discrete action space environments is non-stable, no matter which replay method is used. Footnote 2 and \cite{foerster2016learning,Foerster2017Stabilising} explain this phenomenon in some extent, but further research is needed.

(5) Full-information activation function for sensitive continuous action. Ideally (for continuous action environments), the policy $\pi(a|s)$ is a one-to-one function mapping between state $s$ and optimal action $a^{*}$. If we use a neural network $\pi(a|s;\theta)$ to approximate $\pi(a|s)$ to meet the one-to-one mapping requirement, we should not throw away any (useful) information in state $s$ at any layer of the network $\pi(a|s;\theta)$. Otherwise, similar states may be encoded into identical hidden vector and further be mapped to the same optimal action $a^{*}$. So we suggest to use sigmoid, elu, etc. rather than relu as activation functions for sensitive continuous action space applications. Similarly, we suggest to use relu for discrete action space applications because action is finite and similar states often correspond to the same optimal action $a^{*}$ in those applications. Our preliminary experiments show that relu based models will generate an averagely optimal action but not the exactly optimal action. Further analyses can be found in the supplementary material.

(6) Centralized coordinator. Is there any single point failure? As ACCNet is fully distributed, any agent or special designed agents can act as the coordinator. In addition, the A-CCNet does not need the coordinator during execution, and centralized training is a common setting for MARL systems \cite{foerster2016learning,foerster2017counterfactual,Lowe2017Multi}.

\section{Conclusion}
The proposed ACCNet, born with the combined abilities of deep models and actor-critic reinforce models, is a general framework to learn communication protocols from scratch for fully cooperative, partially observable MARL problems, no matter the action space is continuous or discrete. Specially, the A-CCNet, one concrete implementation of ACCNet, can make the training of MARL systems more stationary than previous methods as supported by both mathematical Equation (11) and experimental results of various environments. Another attractive advantage of A-CCNet is that it does not need communication during execution while still keeps a good generalization ability.

For the future work, we will put our efforts on the following important and challenging problems. (1) How to make the training of discrete action space MARL systems more stationary. Special experience replay method may be a powerful tool for this problem. (2) How to make the communication signals more sparse. In this paper, we use deep neural networks to compress the original communication messages. This method can achieve ``spatial-sparsity'', i.e., the dimension of communication signals is limited and most values are around zero. Another one is ``time-sparsity'', i.e., the communication signals only come \emph{intermittently}. Although those concepts are borrowed from sparse autoencoder \cite{lee2008sparse,makhzani2013k,makhzani2015winner}, they are very useful in the real-world distributed MARL systems. We find that gating mechanism and token mechanism are very useful for achieving ``time-sparsity''. We will introduce our methods more formally in the future.

\subsubsection*{Acknowledgments}
The authors would like to thank Xiangyu Liu, Weichen Ke, Chao Ma, Quanbin Wang, Yiping Song and the anonymous reviewers for their insightful comments. This work was supported by the National Natural Science Foundation of China under Grant No.61572044. The contact author is Zhen Xiao.

\end{document}